# Listen to Your Favorite Melodies with img2Mxml, Producing MusicXML from Sheet Music Image by Measure-based Multimodal Deep Learning-driven Assembly

**Tomoyuki Shishido[1], Fehmiju Fati[2], Daisuke Tokushige[3], and Yasuhiro Ono[4]**

[1]Shishido&Associates, [2]Composer, Ph.D., [3]SK Intellectual Property Law Firm, [4] Enspirea LLC. Correspondence to: Tomoyuki Shishido, Ph.D. shishid@shishidopat.onmicrosoft.com and Yasuhiro Ono yono@joshono.net

**Abstract**

Deep learning has recently been applied to optical music recognition (OMR). However, currently OMR processing from various sheet music images still lacks precision to be widely applicable. Here, we present an MMdA (Measure-based Multimodal deep learning (DL)-driven Assembly) method allowing for end-to-end OMR processing from various images including inclined photo images. Using this method, measures are extracted by a deep learning model, aligned, and resized to be used for inference of given musical symbol components by using multiple deep learning models in sequence or in parallel. Use of each standardized measure enables efficient training of the models and accurate adjustment of five staff lines in each measure. Multiple musical symbol component category models with a small number of feature types can represent a diverse set of notes and other musical symbols including chords. This MMdA method provides a solution to end-to-end OMR processing with precision.

**1. Introduction**

Optical music recognition (OMR) involves computationally reading music notation in sheet music. The challenges to perform OMR with precision remain and have been reviewed elsewhere (Rebelo *et al.*, 2012; Shatri and Fazekas, 2020). One of the goals is to create a machine-readable symbolic format from music scores. The typical architecture of an OMR processing system includes (1) preprocessing, (2) musical symbol recognition, (3) musical notation reconstruction, and (4) final representation construction. Conventional OMR processing systems have utilized classification technology (e.g., support vector machines (SVMs), neural networks (NNs), k-nearest neighbors (kNN), optical character recognition (OCR)) to recognize musical symbols after staff lines removal, while positional information is retained (Rebelo *et al.*, 2012).

## 2. Related Work

To improve precision of the musical symbol recognition, approaches for object detection using deep learning (Zhao *et al.*, 2019) have recently been adopted. Calvo-Zaragoza and colleagues used convolutional neural networks (CNNs) and recurrent neural networks (RNNs) as well as so-called Connectionist Temporal Classification loss function to computationally decode music notation from images (Calvo-Zaragoza *et al.*, 2018). Subsequently, Huang *et al.* used a deep convolutional neural network and feature fusion to directly process the entire image and output musical symbol components with high accuracy (Huang *et al.*, 2019).

Here, we extract measures by using a deep leaning YOLOv5 model (Ultralytics, 2020) while the entire sheet music image is processed; we then adjust and use, in each measure, five staff lines as a positional reference to identify musical symbol components by using multiple YOLOv5 models. That step is followed by assembly of the musical symbol components to reconstruct and digitize the entire sheet music as a MusicXML file, that is, Measure-based Multimodal deep learning (DL)-driven Assembly (MMdA). This MMdA allows even an inclined photo score image to be digitized as a MusicXML file with precision, and can thus provide a solution to end-to-end OMR processing.

## 3. Methods
### 3.1 YOLOv5 Model Training

A YOLOv5 model was cloned from https://github.com/ultralytics/yolov5. To implement transfer learning, an existing pretrained model with yolov5x.yaml was finetuned to create the corresponding models during the following training. The entire score images (.jpg) were used for training of a measure model. In each score image, respective individual bounding boxes were assigned to respective measures by using an annotation tool labelImg (https://github.com/tzutalin/labelImg). The categories for the measures include x0, x1, and y0 (see Fig. 2A for description). The annotated images and text files were formatted for YOLOv5 in Roboflow (https://roboflow.com). The training, validation, and test sets were resized to 416 × 416 pixels, the typical batch size ranged from 16 to 32, and the epochs were from 200 to 2000 depending on the training process.

Basically the same procedure was repeated in order to train musical symbol component feature models by category. However, each measure with musical symbol components was extracted manually from the entire score images, resized to 416 × 416 pixels, annotated using labelImg, and formatted for YOLOv5. The categories of each musical symbol component feature model included accidental, arm/beam, body, clef, and rest feature categories although the categories are incomplete, and are listed in Fig. 4B.

### 3.2 YOLOv5 Model Inference

The weights of trained models were stored and used for inference while the image size was set to 416 × 416 pixels for each input image piece and the confidence rate ranged from 0.60 to 0.90 depending on usage and performance of the model used. Note that in production variants, the input images (either the whole score image or each measure image unit) were leveled horizontally relative to staff lines according to the algorithms described below.

### 3.3 Alignment of Measures

The resulting measures identified and extracted by the measure model were aligned pair-wise (i.e., staff 1 for the right hand or staff 2 for the left hand in a piano music piece). For this purpose, the measures identified as x0 or x1 category (i.e., including a clef) were vertically sorted. Next, each identified measure was used to group and sort horizontally overlapping measures (e.g., y0 measures) in sequence. Misidentified overlapping measures were eliminated in some cases. Then, vertically alternating measure groups were assigned to either staff 1 or staff 2 measure units.

### 3.4 How to Level Five Staff Lines

To horizontally level the entire sheet music image, we adopted a process including: (1) converting an input image into a gray-scale image; (2) extracting edges of the image by using the Canny method (Canny, 1986); (3) detecting straight lines by the Hough transformation (Hough, 1962); (4) calculating the tilt angle of the longest straight line; and (5) rotating the entire input image by the tilt angle calculated.

We further leveled each measure unit image. Substantially the same process described above was carried out except that horizontally extending straight lines were selected for processing.

### 3.5 How to Adjust the Positions of and Spacing Between Five Staff Lines

After staff lines were horizontally leveled, we positioned the staff lines and used them as a positional reference for identification and reconstruction of musical symbol components. We introduced parameters $\alpha$ and $\beta$ for the positioning of the staff lines. $\alpha$ represents a vertical deviation of the middle staff line from the center of the staff. $\beta$ represents an increment/decrement in spacing between the staff lines. These parameters were determined as follows.

(1) The whole vertical width (including the vertical width of the staff and the

lengths of upper and lower extra areas preset (e.g., 1.2 times the vertical length of the staff)) was set to 1.0. The range of α from -0.03 to 0.03 were tested while the value was increased by 0.01. Simultaneously, the range of β was varied between -0.005 to 0.005 while the value was increased by 0.001.

(2) The respective α and β were used to draw and superimpose simulated five lines on the original image.

(3) The resulting image was converted into a gray-scale image and subjected to Gaussian threshold processing to calculate the black area in the processed image.

(4) The α and β when the black area was minimum were determined because the overlapping area between the staff lines in the original image and the superimposed simulated staff lines should be maximum.

(5) The determined α and β were used for correction of the positional reference.

### 3.6 Assembly of Musical Symbol Components

After the inference, each YOLOv5 musical symbol component model gave the type and position of each musical symbol component in the corresponding feature category. This positional information was corrected using the α and β parameters determined. Then, all the information on the musical symbol components were processed and assembled for each measure unit as follows.

(a) Multiple YOLOv5 models for musical symbol components (i.e., accidental, arm/beam, body, clef, and rest categorical features) were applied to each measure unit for inference.

(b) In the measure unit, the above α and β parameters determined were used to map the position of each musical symbol component relative to the staff lines of the measure unit.

(c) All the musical symbol components obtained were aligned horizontally in series.

(d) At that point, we introduced an accidental table that reflected the state of accidental at the current position and could be changed by, for instance, sharp, flat, or natural note. The initial accidental table was set according to the fifths, which represents a key designated by the clef of a music piece of interest. We then updated the status of the clef and the accidental table while each musical symbol component in the measure unit was analyzed horizontally in series. For instance, when the musical symbol component analyzed was a clef type (cf0 or cf1), the status of clef was changed to either G or F clef, respectively. When the musical symbol component was an accidental type (e.g., #, b, ♮), the accidental table was updated according to the accidental type on its position.

(e) When the musical symbol component being analyzed was a rest type, the rest type was identified and put into an output sequence list, except that in the case where the rest type vertically overlapped with a body type and/or an arm/beam type, the rest type was analyzed at the time of annotating the body type and/or the arm/beam type.

(f) When the musical symbol component was a body type, the body type was identified depending on the number and the positions of vertically overlapping feature category types (e.g., an arm/beam type(s), other body type(s), a rest type) aligned in a descending direction (sometimes simply herein referred to as vodMS (vertically overlapping descending Musical Symbols). In the case with a plurality of body types, a chord was assigned to these body types.  The cases for the identification and annotation were examined in the following order.

Case (i): Both the top and bottom feature types were arm/beam types

When the number of body types included in the vodMS was 2, the lower body type was identified in conjunction with the bottom arm/beam type and the upper body type was identified in conjunction with the top arm/beam type while the accidental table at that position was reflected. When there were three or more body types in the vodMS, the distance to the nearest upper body type belonging to the top arm/beam type and the distance to the nearest lower body type belonging to the bottom arm/beam type were calculated for the body type being analyzed in the vodMS. The body type being analyzed was identified in conjunction with the arm/beam type of the nearest body type, whichever was shorter in the distance thereto.  In this case, the body type(s) belonging to the bottom arm/beam type was set to voice 1 and the body type(s) belonging to the top arm/beam type was set to voice 2. The results were put into the output sequence list.

Case (ii): The top feature type was a rest type

The rest type was set to voice 2. One or more body types in the vodMS were identified accordingly and assigned to voice 1. The results were put into the output sequence list.

Case (iii): The bottom feature type was a rest type

The rest type was set to voice 1. One or more body types in the vodMS were identified accordingly and assigned to voice 2. The results were put into the output sequence list.

Case (iv): The top feature type was an arm/beam type

The body type(s) were identified depending on their own type. When the feature type was any of bd0 to bd3, the corresponding note was determined in conjunction with the arm/beam type while the current accidental table was applied.   In the case with a feature type bd4 or bd5 (without an arm (stem)), the corresponding note was provided as it was. Here, the voice was assigned to 1 temporarily and was optionally changed in the voice adjustment step described later.

Case (v): The bottom feature type was an arm/beam type

Basically, the same as in case (iv) was applied to this case except that the bottom arm/beam type was used for the annotation.

Case (iv): Both the top and the bottom feature type(s) were body types

In this case, the body type(s) were assumed to be any of bd4 and/or bd5. However, due to misrecognition of arm/beam types, for instance, the feature body types bd0 to bd3 could be incorporated in the vodMS. In this situation, each body type was identified in conjunction with a probable arm/beam type. The voice of the body types in the vodMS was temporarily assigned to 1 and was optionally changed in the voice adjustment step.

During the annotation of each body type, the clef status and the current accidental table were used, and the corresponding note was identified while the step of the note was calculated relative to the position in the corrected five staff lines.

The rest type arranged at the middle position in the vodMS was excluded from the body type analysis, and the rest type was put into the output sequence list before or after the body type being analyzed, which depends on the position of the rest type relative to the body type.

In addition, the already analyzed body, arm, and rest types were excluded in the subsequent annotation whereas the beam type was reused for the annotation because beam types have start, intermediate, and stop functions.

## 3.7 Voice Adjustment

Each measure has a fixed duration of all the notes combined, which is typically defined by the beats and beat-type as well as the tempo. We checked whether or not the measure of interest had an appropriate duration when note durations from all the notes identified were combined. When the duration was longer than the appropriate duration, the voice of each note was adjusted as follows.

(1) The total duration of notes (derived from body and rest types) assigned to voice 1 or 2 was calculated while a chord was taken into account.

(2) When the total duration for either voice 1 or 2 was longer than a predetermined duration, the following voice adjustment was carried out.

(i) The voice of each body type with the top arm/beam type was changed to voice 2.

(ii) While the voice of each body type with the bottom arm/beam type was set to 1, the remainder (e.g., bd4 or bd5) was changed to voice 2.

(iii) The whole notes (e.g., bd4, bd5) were changed to voice 2.

### 3.8 Generation of the MusicXML File

The output sequence list for each measure unit was assembled as time-series data. For this purpose, the individual notes in each measure unit were structured as a document tree. The resulting tree reflected the entire notation of the original sheet music, and was converted into an XML format for either staff 1 or staff 2 to create a MusicXML file.

This MMdA method and program are currently available, under GPLv3 License, as img2Mxml in the GitHub repository at https://github.com/TomoShishido/img2xml. In addition, a Web application img2Mxml (sheetmusic2MusicXML) is also available at https://ui.saaipf.com/app/upload.

## 4. Experimetal Results

### 4.1 Overview of Measure-based Multimodal DL-driven Assembly (MMdA)

The whole process from an input of sheet music image to production of music information (MusicXML file) is provided in Fig. 1.

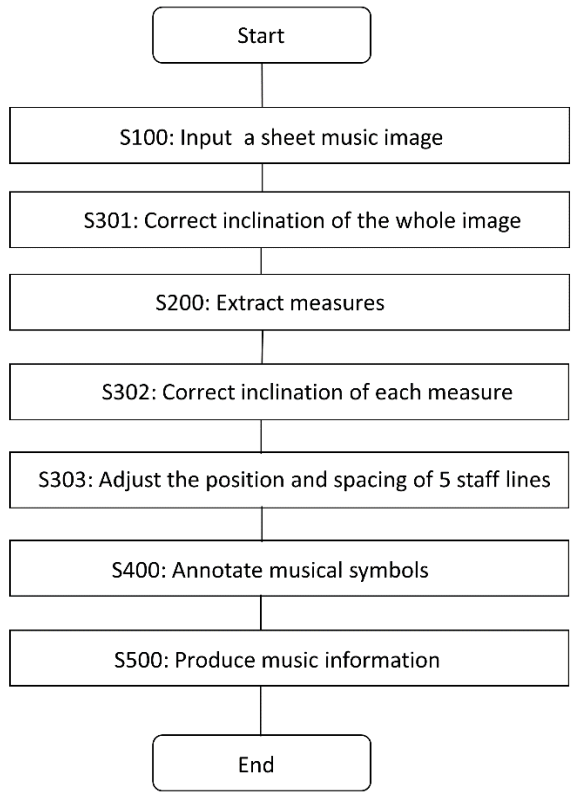

Fig. 1. The whole process: each step (S100, S301, S200, S302, S303, S400, or S500) is indicated and described.

The whole process is briefly described. The input was a sheet music image (S100). Next, the whole sheet music image was leveled horizontally (S301). Then, a measure-recognizing deep learning YOLOv5 was applied to extract measures in the image (S200). The measures were aligned, resized, and divided into staff 1 or staff 2 group. The inclination of each measure was corrected to have an approximately horizontally leveled measure unit (S302). The positions of and spacing between five staff lines in each measure unit were adjusted one by one (S303). Subsequently, five different YOLOv5 models for recognizing given feature types in five feature categories were applied for inference, and the resulting inferred feature types in each feature category were used to annotate musical symbols (e.g., notes with the step of a note) while the musical symbol components were identified/mapped relative to the corrected five staff lines (S400). After that, the annotated musical symbol components were assembled in each measure. Finally, the measures were connected in series to produce a MusicXML file (music information) (S500). This process was named Measure-based Multimodal DL-driven Assembly (MMdA). Each step will be described in detail below.

**4.2 Training of a measure-recognizing deep learning model**

First, we trained the YOLOv5 measure model using 47 full score images (each containing from a few to about 50 measures) and achieved an mAP@0.5 (a metric accuracy) of 0.95. The feature types of measure category in this model included x0, x1, and y0 measure feature types (measures beginning with G clef, F clef, and the other measure types, respectively) as shown in Fig. 2A. To create the training data, we used the labelImg software (https://github.com/tzutalin/labelImg) to assign a bounding box to the corresponding type in each image. The bounding box was set to follow the top and bottom lines of the staff. The training data, test data, and validation data for training were prepared on the website of Roboflow (https://app.roboflow.com/).

**4.3 Inference of measures from a variety of sheet music images (S100 and S200)**

Next, we applied this measure model to inference for a score image that was not used for training. Fig. 2B shows the inference results of a PDF-derived image obtained by scanning a portion of the score of Handel's "Sarabande and Variation". Fig. 2C shows the inference results of a photo image of the same score as obtained using a smartphone camera. As a result, 100% of the measures in each score image were recognized and extracted with an inference confidence rate of 0.91 to 0.95. Part of the score of Beethoven's "Pathetique 2nd Movement" (used for training this measure model) was also recognized, and all the measure were extracted with an inference confidence rate of 0.92 to 0.93 (data not shown).

Further, for the score image of Bach's Menuet (not used for training this measure model), the confidence rate of the inference ranged from 0.79 to 0.93, and overall about 94% of the measures were correctly recognized. However, 1 measure out of 66 measures was recognized as both x0 and x1, and 2 measures were fused together. In addition, one measure contained one adjacent note. The results are shown in Fig. 2D.

The results have demonstrated that this measure model is useful for efficiently extracting measures in PDF-derived or photo images of musical scores that were not used for training.

A.

| Type of Measure | Description |
| --- | --- |
| x0 | Measure starting with a G clef |
| x1 | Measure starting with an F clef |
| y0 | The remainder |

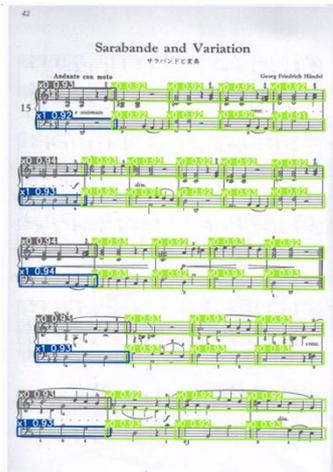
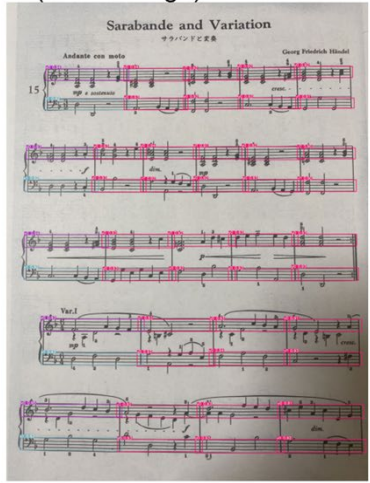
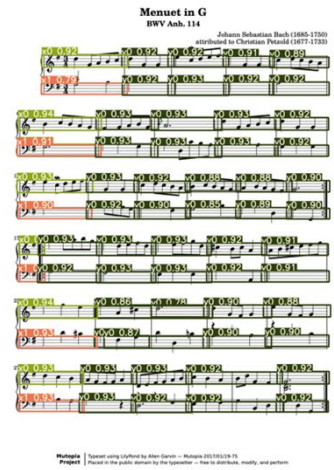

B (PDF-derived Image)   C (Photo Image)   D

Fig. 2. (A) Types of measure recognized by a YOLOv5 deep learning model. The results of inference were shown for a PDF-derived image (B) or a photo image (C) of Sarabande and Variation by Handel, and a PDF-derived image of Menuet by Bach (D).

**4.4 How to Level Inclined Image and Each Measure (S301 and S302)**

Fig. 3A shows an inclined photo image of a Sarabande score. Since the staves do not function as a positional reference unless they are in a horizontal state, the entire photo image was leveled horizontally in accordance with the protocol described in the Method Section 3.4 (Fig. 3B). After the measures were extracted with the above measure model, some of the measures were still inclined. Thus, each measure was further likewise leveled horizontally (Fig. 3C).

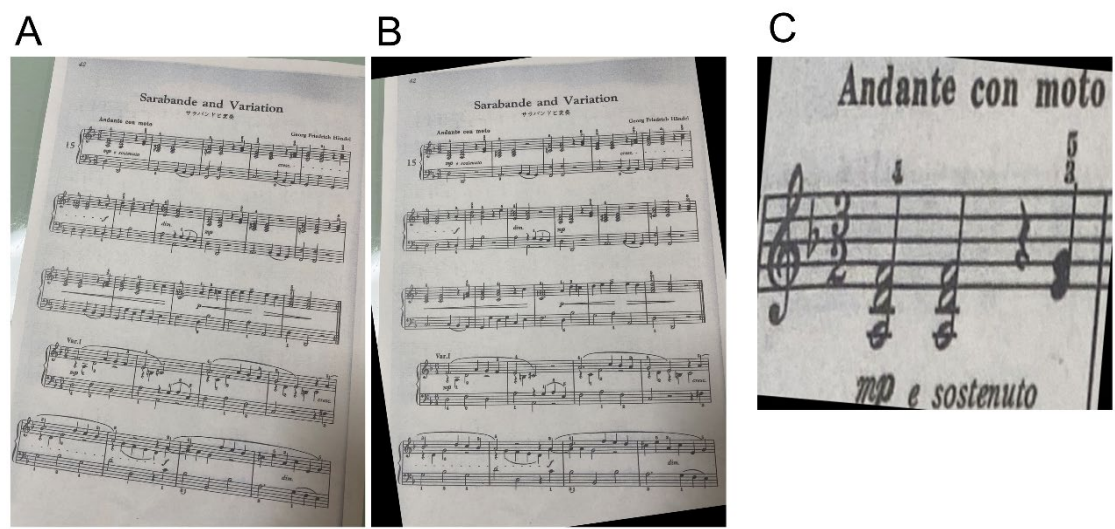

Fig. 3. (A) An inclined photo image of Sarabande and Variation. (B) A horizontally leveled image of (A). (C) A horizontally leveled measure among measures extracted by a YOLOv5 measure model.

**4.5 Training and Inference of YOLOv5 Models for Musical Symbol Components in Each Measure**

The feature categories used for the models included accidental, arm/beam, body, clef, and rest categories (Fig. 4A), and the number of the feature types in each category was 3 (ac0, ac1, ac2), 8 (am0, am1, am2, am3, bm0, bm1, bm2, bm3), 6 (bd0, bd1, bd2, bd3, bd4, bd5), 5 (cf0, cf1, cf2), and 5 (re0, re1, re2, re3, re4, re5), respectively (see Fig. 4B for description).

Individual measures, in which feature types of each feature category had been identified, were used to train each feature category model.  The number of training images (measures) was 199 for accidental, 546 for arm/beam, 537 for body, 149 for clef, or 611 for rest.  The mAP@0.5 was about 0.94 or higher for each feature category. Although the number of training data images was low, the metric mAP@0.5 was high. This may be because we used transfer learning and each measure was standardized in terms of input size (see Method Section 3.1).

The results of inference using each feature category model are shown in Fig. 4C. The results demonstrated that each feature category model trained was sufficient for recognition of the corresponding feature types of the feature category in each measure tested whereas the confidence rate of each feature type varied. Thus, the confidence threshold of inference should be set depending on each feature category model trained.

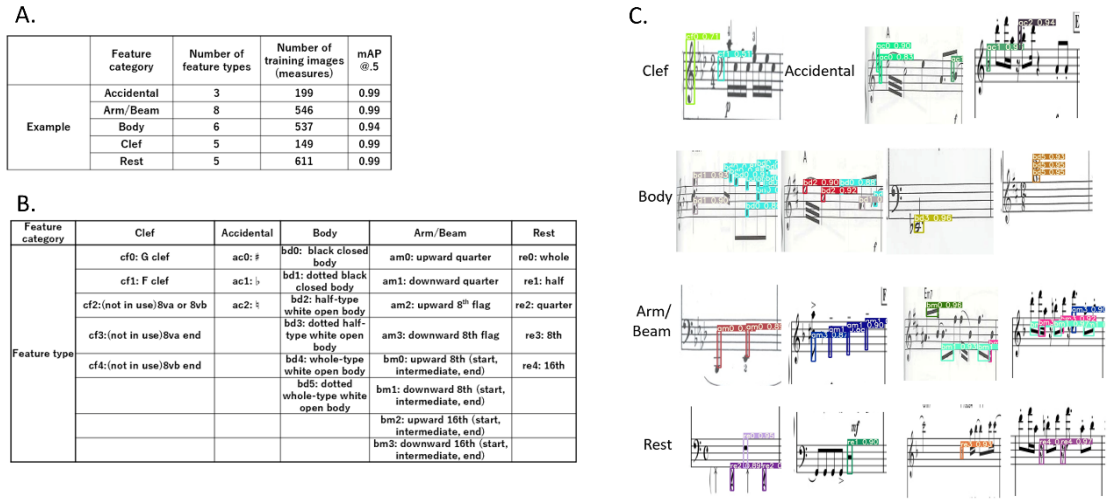

Fig. 4. Information about (A) the feature categories and training and (B) their feature types. (C) The results of inference by each feature category model were provided.

**4.6 Parallel or Sequential Processing Time (CPU vs. GPU)**

Since we used five different feature category models for inference of musical symbol components in each measure, we measured the time required to process sheet music image data in parallel or in sequence.

Specifically, using the deep learning measure model created, we processed each sheet music image to obtain measures. Next, the measures were aligned and resized. Then, we applied the above five musical symbol component models to each measure. This process of creating analysis data was automated, and the time taken for the process was measured. We compared the processing time when the five musical symbol component models were processed in sequence or in parallel. The results are shown in Fig. 5.

We used three different sheet music images including Menuet with 66 measures, Sarabande with 48 measures, and Pathetique 2nd Mov. with 58 measures. The experiments were conducted with an iMacPro (CPU processor: 3.2 GHz, 8-core Intel Xeon W; Memory: 64 GB 2666 MHz DDR4), and were triplicate and averaged (Fig. 5A). The average time for sequential processing was 153.8 sec, 121.5 sec, and 138.1 sec for

Menuet (66 measures), Sarabande (48 measures), and Pathetique 2nd movement (58 measures), respectively. The time was almost proportional to the number of measures. The average time for parallel processing was 81.3 sec, 63.0 sec, and 75.4 sec for Menuet, Sarabande, and Pathetique, respectively. This time was also almost proportional to the number of measures. By parallel processing, the processing time of Menuet, Sarabande, or Pathetique was reduced by about half to 52.9%, 51.9%, or 54.6%, respectively.

Next, the experiments were conducted using GPUs in AWS EC2 instance g4dn.metal (Fig. 5B). The CPU/GPU configuration of g4dn.metal included 8 NVIDIA T4 Tensor Core GPUs, 96 vCPUs, and 384 GiB RAM, etc. The GPUs were processed in sequence or parallel while the same automated process above was used. The average processing time for the Menuet score in the case of using GPUs in sequence was 70.9 seconds, which was shorter than the average processing time of 153.8 sec obtained when the CPU was used in sequence and 81.3 seconds when the CPU was used in parallel. The parallel processing time was 16.4 sec on average, which was about 1/4 of the sequential processing time. This GPU parallel processing time was about one-tenth of the CPU sequential processing time, demonstrating that GPU parallel processing should significantly reduce processing time.

### A

| Sheet music | Processing | Time (sec) | | | | Parallel/Sequential (%) |
|---|---|---|---|---|---|---|
| | | 1st | 2nd | 3rd | Average | |
| Menuet (with 66 measures) | Sequential | 160.8 | 150.4 | 150.2 | 153.8 | 52.9 |
| | Parallel | 79.8 | 82.6 | 81.5 | 81.3 | |
| Sarabande (with 48 measures) | Sequential | 120.7 | 123.8 | 120.0 | 121.5 | 51.9 |
| | Parallel | 61.5 | 61.6 | 65.9 | 63.0 | |
| Pathetique 2nd Mov. (with 58 measures) | Sequential | 138.5 | 138.9 | 137.0 | 138.1 | 54.6 |
| | Parallel | 81.8 | 72.8 | 71.7 | 75.4 | |

### B

| Sheet music | Processing | Time (sec) | | | | Parallel/Sequential (%) |
|---|---|---|---|---|---|---|
| | | 1st | 2nd | 3rd | Average | |
| Menuet (with 66 measures) | Sequential | 71.1 | 70.8 | 70.9 | 70.9 | 23.1 |
| | Parallel | 16.6 | 16.2 | 16.3 | 16.4 | |

Fig. 5. (A) The processing time when a CPU with 8 cores was used in sequence or in parallel. (B) The processing time when GPUs were used in sequence or in parallel.

**4.7 Adjustment of the Positions of and Spacing between Five Staff Lines Before**

**Annotation (S303)**

As described in Method Section 3.5, the positions and spacing of five staff lines in each measure were adjusted using parameters α and β. The initial calculated five staff lines were deviated from the real ones (Fig. 6A). Then, the parameters α and β were programmatically determined, and used for the adjustment. The calculated five staff lines adjusted were fit for the real ones (Fig. 6B). In this way, the positions of five staff lines adjusted were used as a positional reference for the following identification and annotation.

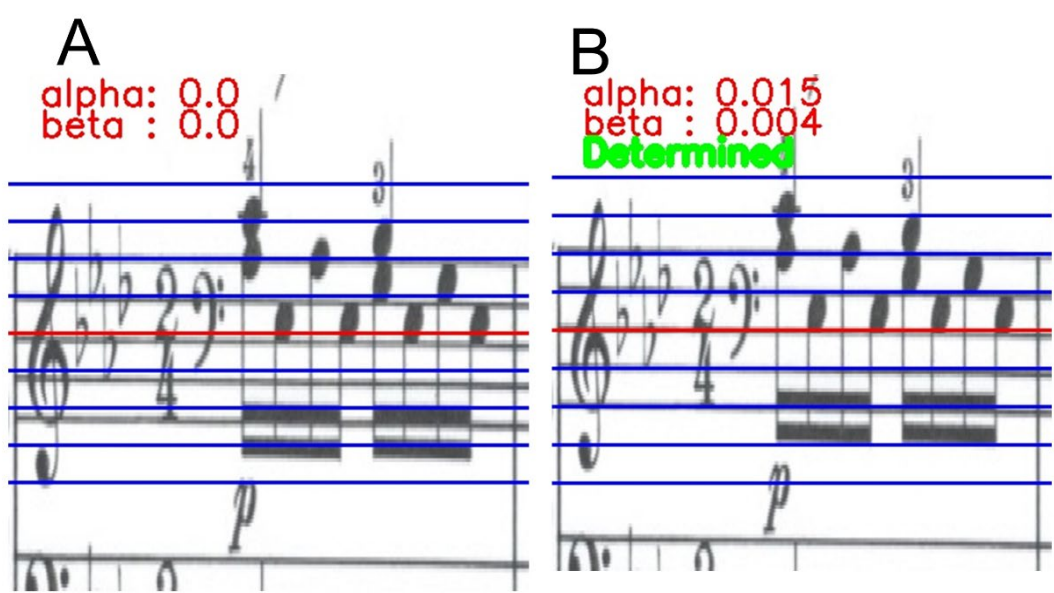

Fig. 6. (A) The calculated staff lines before adjustment using parameters α and β. (B) The simulated staff lines after the adjustment. A red line indicates the middle staff line, and blue lines indicate the other five staff lines and other corresponding lines in extended staff areas.

## 4.8 Production of MusicXML by Identifying Musical Symbol Components and Annotating Notes, etc., by MMdA (S400 and S500)

Analysis data obtained through the above process was analyzed using the protocols described in Method Sections 3.6 to 3.7 to identify musical symbol components and annotate notes, etc., by MMdA.

The sheet music image of Fig. 2D (Menuet by Bach) was used to identify musical symbol components and annotate notes, etc., as described above. Then, a MusicXML file for each staff was produced by the protocol described in Method Section 3.8. The resulting MusicXML file for staff 1 was successfully incorporated into music notation software Sibelius (Fig. 7A) or MuseScore (Fig. 7B).

For staff 1, measures were recognized with 97% (32/33) accuracy, demonstrating the high accuracy of measure extraction. For musical symbols annotated by identifying and combining the individual feature types with a staff line positional reference: step (based on a clef-type and the positional reference), note (also including the duration), and chord (having a perfect match), the accuracy was 98% (125/128), 95% (122/128), and 100% (1/1), respectively. Accidental symbols were also recognized with 100% (3/3) accuracy (Fig. 7E, the third row).

For staff 2, measures were recognized with 97% (32/33) accuracy. For the step, note, and chord, the accuracy was 95% (71/75), 95% (71/75), and 100% (1/1), respectively. Rest symbols were recognized with 40% (2/5) accuracy. The accuracy of accidental was 50% (1/2) (Fig. 7E, the fourth row).

These results show that the accuracy of the notes annotated by this method is high.

In addition, we examined whether MusicXML could be created not only from PDF images, but also from photo images, which are likely to be used in practice. In this case, the staves of each photo image are often not horizontal. Thus, a MusicXML file was created, using the protocols in the Method Sections, from an inclined photo image of Sarabande and Variation shown in Figure 7C. The resulting MusicXML file was successfully incorporated into Sibelius (Fig. 7D).

As shown in Fig. 7E, measures were recognized with an accuracy of 96% (23/24). For the step, note, and chord, the accuracy was 87% (135/156), 86% (134/156), and 78% (29/37), respectively. Rest symbols were recognized with 64% (16/25) accuracy. The accuracy of accidental was 71% (10/14). In particular, the Sarabande score contained 37 relatively complex chords in staff 1. These chords were recognized with an accuracy of 78%.

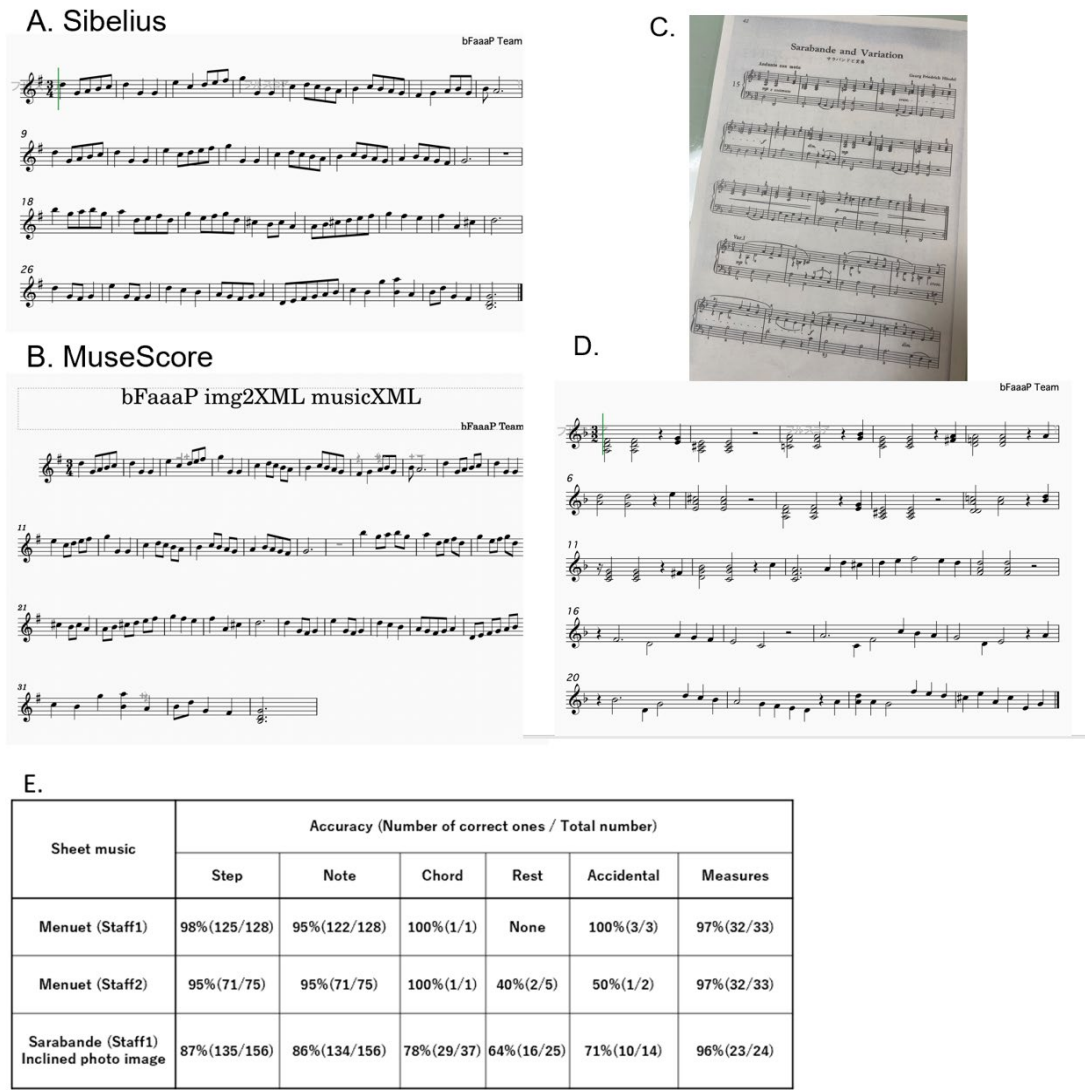

Fig. 7. A MusicXML file created from a Bach's Menuet score image by the MMdA method was incorporated into music notation software Sibelius (A) or MuseScore (B). An inclined photo image of Sarabande and Variation by Handel (C) was converted by the MMdA method to a MusicXML, which was incorporated into Sebelius (D).   The results of accuracy of music notation by the MMdA method were listed (E).

## 5. Conclusion

Here, the MMdA method has been used to demonstrate that MusicXML files can be produced with precision from various sheet music images. We discuss the advantages and disadvantages of the MMdA method as follows.

### 5.1 Measure-based Approach

**Advantages**

(i) To facilitate training of each deep learning model

As described in the method and result sections, a relatively small number of training images can be used to carry out transfer learning for each musical symbol component deep learning model. Because each training measure image was resized to a fixed area (416 x 416 pixels), this standardization of the size of input image data seemed to facilitate the training.   This makes it possible to further train the models with new feature types or even another model with a new feature category efficiently.

(ii) Efficient adjustment of the positions of and spacing between five staff lines

Since each measure was used to adjust the positions of and spacing between five staff lines, the positions and spacing were more accurately determined. This can improve identification and annotation of notes with correct steps, which are each determined using a clef type and a position relative to the adjusted five staff lines.   In regular photo images of sheet music, the staves of each image are usually variously inclined.   Thus, just the overall horizontal leveling of the image is insufficient, and leveling of the staff lines in each measure gives a better adjustment (see Figs. 3 and 6).

## 5.2 Multimodal DL-driven Approach

**Advantages**

(i) To annotate diverse musical symbols by using a smaller number of feature categories and types

For G clef, we assigned 25 steps from D3 to G6 on the basis of the position of the staff. For F clef, we assigned 25 steps from F1 to B4. Depending on the position of the body type, the total of 2 (G and F clefs) x 25 (steps) x 6 (the number of body types) = 300 different variations can be identified. In addition, the duration of each note is determined by the arm/beam type (whole notes do not take arm/beam, and half notes only take am0 or am1). The three beam types (start, intermediate, and end) indicate the position in the beam. Therefore, 300 x 2 (2 types of whole notes) + 300 x 2 (2 types of half notes) x 2 (am0 and am1) + 300 x 2 (black closed body types) x (4 (arm types) + 4 (beam types) x 3 (start, intermediate, and end)) = 11,400. Also, there are 3 different accidentals to modify the 11,400 notes. Thus, from the 19 feature types, about 30,000 new notes with a step and an accidental can be expressed. Furthermore, we can take chords into consideration. Since a chord is any combination of 2, 3, 4, and 5 notes, the number of feature types that can be represented increases dramatically, and we can easily represent more than 100,000 types of single notes and chords. Hence, we have demonstrated that a large number of new note feature types can be identified and annotated by combining a relatively small number of feature types in multiple categories.

(ii) Chords can be identified

The 6 different body types were used. When the multiple bodies with possibly different types are vertically aligned during analysis, various chords can be assigned.  As shown in Figs. 7D and E, relatively complex chords were annotated with an accuracy of 78%. Thus, the MMdA method allows for notation of chords.

**Disadvantages**

(i) Long inference processing time

Since we used one measure-recognizing deep learning model and five different, measure-based, musical symbol component deep learning models, the entire inference processing time was long (see Fig. 5). As the number of measures used for inference increases, the processing time becomes longer and is proportional to the number of measures.  Also, in order to provide a complete set of musical symbols in notation, we need more feature categories and feature types.  This also further increases processing time.

### 5.3 CPU vs. GPU settings

To reduce the inference processing time, we examined the sequential or parallel processing using a CPU(s) or GPUs (Fig. 5). Although the number of GPUs used was not proportional to the reduced processing time, the GPU parallel processing dramatically decreased the processing time to one-tenth of the CPU sequential processing time.  This suggests that multiple GPUs used in parallel for the processing should decrease the overall processing time to produce a final MusicXML file more quickly. However, the OMR processing at this purpose is not necessarily performed in real time.

### 5.4 Future Perspectives

In the case of handwritten score images, the musical symbol component deep learning models can be further trained using measure image data with handwritten musical symbols. This will extend the application to various handwritten musical scores.

Meanwhile, the actual MusicXML files (Menuet by Bach and Sarabande and Variation by Handel) are currently provided at bfaaap/musicdata folder of open source project img2Mxml at https://github.com/TomoShishido/img2xml. When an actual music sound was produced using conventional music notation software (e.g., Logic Pro, MuseScore, Sibelius), the 78% accuracy of chords in the MusicXML for staff 1 of Sarabande was not perfect, but favorable.  Thus, when the notation is corrected, the efforts are less than in conventional methods.

The accuracy of notation will be further enhanced by more data generated by additional training via the deep learning model. Although the current subject is intended to be used for pieces of piano sheet music, more complex scores including many parts may be processed by revising the analysis algorithm. This can extend use of this MMdA method to a variety of scores including those for an orchestra.

In addition, GPU settings will be improved to have multiple GPUs available even for inference. We can take advantage of the improved GPU settings to reduce the inference processing time during MMdA.

Further, the process of identification and annotation by MMdA may be also subject to machine learning such as XGBoost (Chen and Guestrin, 2016) or deep learning. This automation may be used to minimize the complexity of assembly of musical symbol components inferred by the current musical symbol component models.

In conclusion, we have presented a new method of OMR, MMdA. This MMdA method can be used to produce a MusicXML file with precision, and can thus provide a solution to end-to-end OMR processing using deep learning models.

**Acknowledgements**

We thank Dr. Gunnar Dietz for critically reading the manuscript. This work is part of bFaaaP project, and has been funded by Shishido&Associates.

**Conflict of interest statement**

Declaration of interests: T.S., F.F., D.T., and Y.O. are inventors on a patent application covering this method.